# Language Models sounds the Death Knell of Knowledge Graphs

Kunal Suri
Optum, India
kunal_suri@optum.com

Atul Singh
Optum, India
atul_singh18@optum.com

Prakhar Mishra
Optum, India
prakhar_mishra29@optum.com

Swapna Sourav Rout
Optum, India
rout.swapnasourav@optum.com

Rajesh Sabapathy
Optum,India
rajesh_sabapathy@uhc.com

*Abstract*—Healthcare domain generates a lot of unstructured and semi-structured text. Natural Language processing (NLP) has been used extensively to process this data. Deep Learning based NLP especially Large Language Models (LLMs) such as BERT have found broad acceptance and are used extensively for many applications. A Language Model is a probability distribution over a word sequence. Self-supervised Learning on a large corpus of data automatically generates deep learning-based language models. BioBERT and Med-BERT are language models pre-trained for the healthcare domain. Healthcare uses typical NLP tasks such as question answering, information extraction, named entity recognition, and search to simplify and improve processes. However, to ensure robust application of the results, NLP practitioners need to normalize and standardize them. One of the main ways of achieving normalization and standardization is the use of Knowledge Graphs. A Knowledge Graph captures concepts and their relationships for a specific domain, but their creation is time-consuming and requires manual intervention from domain experts, which can prove expensive. SNOMED CT (Systematized Nomenclature of Medicine - Clinical Terms), Unified Medical Language System (UMLS), and Gene Ontology (GO) are popular ontologies from the healthcare domain. SNOMED CT and UMLS capture concepts such as disease, symptoms and diagnosis and GO is the world's largest source of information on the functions of genes. Healthcare has been dealing with an explosion in information about different types of drugs, diseases, and procedures. This paper argues that using Knowledge Graphs is not the best solution for solving problems in this domain. We present experiments using LLMs for the healthcare domain to demonstrate that language models provide the same functionality as knowledge graphs, thereby making knowledge graphs redundant.

## I. Introduction

Knowledge graphs (KG) are knowledge bases that capture concepts and their relationships for a specific domain using a graph-structured data model. Systematized Nomenclature of Medicine – Clinical Terms (SNOMED CT) (SNOMED), Unified Medical Language Systems(UMLS) [Bodenreider O. 2004], etc., are some of the popular KG in the healthcare domain. Fig. 1 shows a sample from a representative medical entity, KG. On the other hand, a language model is a probability distribution over a word sequence and is the backbone of modern natural language processing (NLP). Language models try to capture any language's linguistic intuition and writing, and large language models like BERT [Devlin et al., 2019] and GPT-2 [Radford et al., 2019] have shown remarkable performance. The paper presents a study demonstrating that language models' ability to learn relationships among different entities makes knowledge graphs redundant for many applications.

This paper uses similar terms from SNOMED-CT KG and passes them through a language model for the healthcare domain BioRedditBERT to get a 768-dimensional dense vector representation. The paper presents the results for analyzing these embeddings. The experiments presented in the paper validate that similar terms cluster together. The paper uses simple heuristics to assign names to clusters. The results show that the cluster names match the names in the KG. Finally, the experiments demonstrate that the cosine similarity of vector representation of similar terms is high and vice versa.

Our contributions include: (i) We propose a study to demonstrate the value and application of Large Language Models (LLMs) in comparison to Knowledge Graph-based approaches for the task of synonym extraction. (ii) We extensively evaluate our approach on a standard, widely accepted dataset, and the results are encouraging.

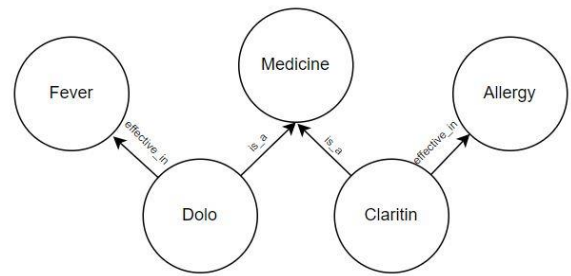

Fig 1. Medical entity Knowledge Graph Representation

The rest of the paper is organized as follows: Section II presents the background required to understand the work presented in this paper. Section III presents a literature survey of related work on knowledge graphs and language models. Section IV presents our understanding of how current days language models are making knowledge graphs redundant. Section V describes our proposed approach. Section VI describes the experiments conducted and the results obtained. Finally, section VII summarizes our work and discusses possible directions for future study.

## II. BACKGROUND

This section defines and describes Language Models and Knowledge Graphs as used in this paper:

*A. Language Models*

A Language Model predicts the probability of a sequence of words in a human language such as English. In the equation below P(w1,…wm) is the probability of the word sequence S, where S = (w1, w2, …, wm) and wi is the ith word in the sequence.

$$P(w_1, \ldots, w_m) = \prod_{i=1}^{m} P(w_i \mid w_1, \ldots, w_{i-1})$$

Large Language Models (LLMs) are language models trained on large general corpora that learn associations and relationships among different word entities in an unsupervised manner. Large Language Models (LLMs) are considered universal language learners. LLMs such as BERT and GPTare deep neural networks based on transformer architecture. One of many reasons for the immense popularity of LLMs is that these models are pre-trained self-supervised models and can be adapted or fine-tuned to cater to a wide range of NLP tasks. Few-shot learning has enabled these LLMs to be adapted to a given NLP task using fewer training samples.

Another reason for the immense popularity of LLMs is that a single language model is applicable for multiple downstream applications such as Token classification, Text classification, and Question answering. LLMs generate embeddings or word vectors for words, and these embeddings capture the context of the word in the corpus. This ability of LLMs to generate embeddings based on the corpus makes them ubiquitous in almost NLP tasks.

In this paper, we use BioRedditBERT [Basaldella et al., 2020], a variant of BERT trained for the healthcare domain. It is a domain-specific language representation model trained on large-scale biomedical corpora from Reddit.

*B. Knowledge Graphs*

Knowledge Graphs (KGs) organize data and capture relationships between different entities for a domain. Domain experts create KGs to map domain-based relations between various entities.

Knowledge graphs are Graph data structures with nodes and edges. Nodes or vertices represent entities of interest, and edges represent relations between them, as shown in Fig 1. KGs can map and model direct and latent relationships between entities of interest. Typically, KGs are used to model and map information from model sources. Once KGs are designed, typically, NLP is used to populate & create the knowledge base from unstructured text corpora.

Knowledge graphs play a crucial role in healthcare knowledge representation. There are many widely used knowledge graphs like SNOMED and UMLS etc. In healthcare, KGs are used for drug discovery drugs, identifying tertiary symptoms for diseases and augmented decision-making, etc.

COMETA: A Corpus for Medical Entity Linking in social media [Basaldella et al., 2020] – a corpus containing four years of content in 68 health-themed subreddits and annotating the most frequent with their corresponding SNOMED-CT entities. In this paper, we have used COMETA to obtain synonyms from SNOMED-CT.

III. RELATED WORK

In 2019, Jawahar et al. performed experiments to understand the underlying language structure learned by a language model like BERT [Ganesh Jawahar et al. 2019]. The authors show that BERT captures the semantic information from the language hierarchically through experiments. BERT captures surface features in the bottom layer, syntactic elements in the middle and semantic features in the top layer. The work presented in this paper treats the BERT model as a black box and demonstrates that BERT can learn the information in a knowledge graph through experiments on real-life healthcare use cases.

There have been studies to generate a knowledge graph directly from the output of LLMs. [Wang C et al., 2020; Wang X et al. 2022] proposes a mechanism to create a KG directly from LLMs. This mechanism talks about a two-step mechanism to generate a KG from LLM. In the first step, different candidate triplets are created from the text corpus. Attention weights from a pre-trained LLM are used to get the best-matched candidate triplets and then validated through a beam search. In the second stage, the matched candidate triplets are mapped to a pre-defined KG for validation, and the unmatched candidates are used to create an open knowledge graph. The work demonstrates the feasibility of the idea presented in this paper that LLM can be used as a substitute for knowledge graphs, especially since they contain the information in the KG.

There is a body of research on integrating Knowledge graphs and LLMs. Structured knowledge from Knowledge Graphs is effectively integrated into Language models to enhance the pre-trained language models [Lei He et al., 2021]. However, these approaches have found limited success, thereby strengthening the position in this paper that LLMs contain information from KGs.

IV. LANGUAGE MODELS FOR KNOWLEDGE GRAPHS

Language Models can find associations between different words based on the attention weight matrix. The methodology to use attention weights as a measure of relationship among the entities indicates that Knowledge graphs are getting replaced by LLMs as they learn more generic relationships in an unsupervised way. The proposed methodology in this paper is built on this idea to demonstrate that Knowledge graphs are increasingly getting redundant for many NLP tasks.

V. PROPOSED APPROACH

The paper demonstrates that language models' ability to learn relationships among different entities makes knowledge graphs redundant for many applications. To illustrate this, we have used word embeddings for all the synonyms of a set of medical terms from a large language model. This work uses

COMETA data to obtain synonyms for a set of medical terms. In COMETA data, the work focuses on the following columns: a) Example column, which contains the sentences from health-themed forums on Reddit, b) Term column contains the medical terms present in the Example column, c) General SNOMED Label column; contains the literal meaning of the Term column from the SNOMED Knowledge Graphs. To obtain synonyms, we use the different values from the Terms column for a specific value of the General SNOMED Label column. For example, for Abdominal Wind Pain General SNOMED label, we have the following three synonyms that we can obtain from the Terms column: gas pains, painful gas, and gas pain.

To calculate the word embeddings of every synonym term, we use the word_vector function from biobert_embeddings python module [Jitendra Jangid, 2020]. Since the original code was incompatible with the current version of Pytorch [Paszke, A. et al., 2019] and Huggingface [Wolf et al., 2020], we modified it just enough to satisfy the current version requirements – the core logic remains the same. We tokenize every Term using HuggingFace tokenizers and pass the tokenized Term through BioRedditBERT model. The previous step gives us embedding for the Term (or sub-terms if the model didn't see the Term before). If the model has not seen the Term before, then we sum up the embedding of all the subterms). We then store all the embeddings for the next steps.

We perform the following two experiments after generating the word embeddings for the synonyms of a set of medical terms. In the first experiment, we cluster the word embeddings for the synonyms of a set of medical terms and assign names to clusters. The word embeddings are passed into UMAP to generate a 2-dimensional representation. We plot the 2-dimensional representation to examine how the term cluster visually. UMAP is used as the dimensionality reduction technique over PCA because it is a non-linear dimensionality reduction technique and does very well to preserve the local and global structure of the data as compared to PCA. However, unlike PCA [Karl Pearson F.R.S. , 1901], UMAP is very sensitive to hyperparameters that we chose, so we visualize the embeddings for several values of number of neighbours (n_neighbors) and minimum distance (min_dist). This step will help us visually validate that a fine-tuned LLM indeed groups together similar terms while ensuring different terms are further apart.

After identifying clusters from the above step, we use Humans in the Loop approach to identify all terms that belong together and run KMeans Clustering Algorithm [Lloyd, Stuart P., 1982] on them. We identify the term closest to the cluster's centroid, which becomes the Parent Node – one of the core uses of Knowledge Graphs.

In the second experiment, we analyze the similarity between the word embeddings of the synonyms of the set of medical terms. In this step, we compute the cosine similarity between all the word embeddings and then we examine the similarity to demonstrate that the synonyms for the same term are similar with a small cosine distance between them.

## VI. EXPERIMENTS AND RESULTS

We use *Term* and *General SNOMED Label* columns from COMETA dataset for our experiments. To calculate the embeddings of every term, we use *word_vector* function from biobert_embeddings package [Jitendra Jangid, 2020]. Since the original code was incompatible with current version of Pytorch [Paszke, A. et al., 2019] and Huggingface [Wolf et al., 2020], we modified it just enough to satisfy the current version requirements – the core logic remains the same.

To test the rich representation of language models for our use case, we perform 2 experiments, (1) Cluster the word embeddings for the synonyms of a set of medical terms and assign names to clusters (2) Analyze the similarity between the word embeddings of the synonyms of the set of medical terms.

For the reasons discussed in Sec. III, we use UMAP as our choice of dimensionality reduction. For experiment (1), Fig. 2 shows that entities having similar nature are grouped together and dissimilar entities are further apart which proves utility of a Fine-tuned Language Models.

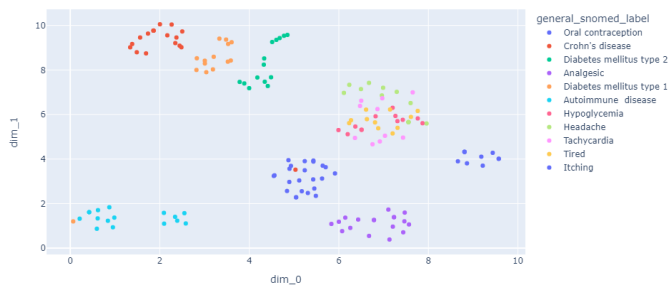

Fig 2. Clusters resulting from UMAP dimensionality reduction

Next we perform KMeans clustering on mentions belonging to same group using cosine similarity. The centroid of each clusters were then used to identify concepts by finding terms that were closest to the centers by cosine similarity. We found the following terms for the concepts visible in Table. 1.

| Concept (General SNOMED Label) | Term (closest to the cluster) |
|---|---|
| Oral contraception | hormonal BC pills |
| Crohn's disease | crohns disease |
| Diabetes mellitus type 2 | T2 diabetes |
| Analgesic | Pain Medication |
| Diabetes mellitus type 1 | T1 diabetic |
| Autoimmune disease | autoimmune disease |
| Hypoglycemia | low blood sugars |
| Headache | head pain |
| Tachycardia | heart racing |
| **Tired** | feel tired |
| **Itching** | itching |

Table 1. Terms closest to the cluster center of each Concept

While Fig. 2 illustrates global and local structure among different mentions of a concept, as a part of experiment (2), we also analyze distribution of similarity scores (which are calculated by using cosine similarity) to visualize distribution of cosine similarity among terms belonging to same concept (Fig. 3 and 4) and terms belonging to different concepts (Fig. 5). We can see that distribution of mentions belonging to same concept are closer to each other on average as compared to mentions from different concepts. This point again validates the utility of Language Model in finding different mentions of a concept in multiple documents.

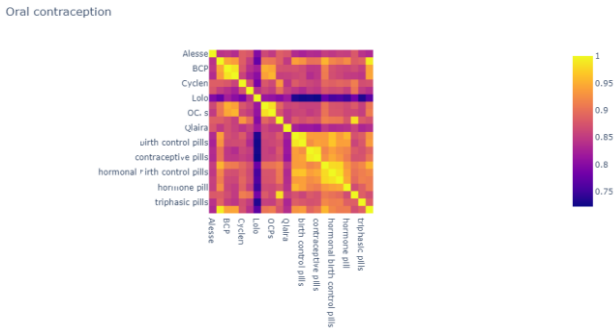

Fig 3. Cosine similarity between mentions from Oral Contraception

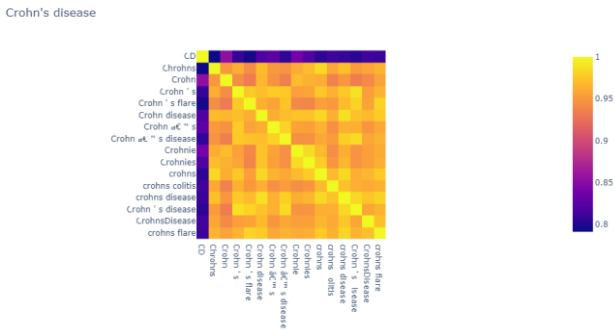

Fig 4. Cosine similarity between mentions from Cron's disease

In addition to these plots, we also analyze similarity between unrelated terms, and we see the following trend –

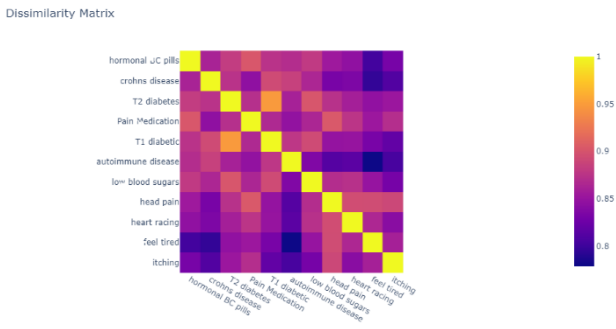

Fig 5. Cosine similarity between mentions from different concepts

## VII. CONCLUSION AND FUTURE WORK

In this paper we have empirically shown how Language Models fine-tuned on domain specific data can be used to replace Knowledge Graphs for tasks where identifying synonyms is involved.

Language Models do a very good job in calculating embeddings which contains semantic information about terms that can be used to identify if two terms are close to each other or not. This information is used in this paper to identify terms which are closer to each other, and which are not. Once groups of similar terms have been identifying using non-linear dimensionality techniques, using Humans in the Loop approach we can annotate such groups. After annotating the groups, we use KMeans to identify centroids of each cluster which are then used the identify terms with the closest cosine distance from them. These terms can then be used as parent nodes for their respective clusters. The primary way in which our algorithm improves over current Knowledge Graph based approaches is that unlike KGs which are created by subject matter experts, our algorithm doesn't require subject matter experts for annotation.

Our current algorithm handles synonym mapping quite well, but it requires human intervention and for next steps, we would be exploring ways in which we can extract Knowledge Graphs from Language Models themselves. This would be required to remove the human intervention in the current process and handling cases where hypernyms are involved.